\documentclass{article}

\usepackage{microtype}
\usepackage{graphicx}
\usepackage{svg} % for including SVG logos via \includesvg
\usepackage{subfigure}
\usepackage{booktabs} % for professional tables
\usepackage{amsmath}

\usepackage{hyperref}

\usepackage[accepted]{subconscious}

\mlsystitlerunning{Adversarial Speculative Reasoning Improves Long Horizon Agents}

\begin{document}

\twocolumn[
\mlsystitle{Adversarial Speculative Reasoning \\Improves Long Horizon Agents}

\mlsyssetsymbol{equal}{*}

\begin{mlsysauthorlist}
\mlsysauthor{Hongyin Luo}{equal,subc,mit}
\mlsysauthor{Wei Fang}{equal,subc}
\mlsysauthor{Lijie Yang}{prin}
\mlsysauthor{Owen Stepan}{subc,neu} \\
\mlsysauthor{Daniel Suit}{subc,neu}
\mlsysauthor{Jeff Krapf}{subc,neu}
\mlsysauthor{Dana Wensberg}{subc}
\mlsysauthor{Jack O'Brien}{subc}
\end{mlsysauthorlist}

\mlsysaffiliation{subc}{Subconscious Systems Technologies, Inc.}
\mlsysaffiliation{mit}{Massachusetts Institute of Technology}
\mlsysaffiliation{prin}{Princeton University}
\mlsysaffiliation{neu}{Northeastern University}

\mlsyscorrespondingauthor{Hongyin Luo}{hongyin@subconscious.dev}

% Affiliations + contact info inline (instead of in the first-page footnote)
\vskip 0.08in
{\centering
\printAffiliationsInline{\mlsysEqualContribution}\par
}

\mlsyskeywords{Machine Learning, MLSys}

\vskip 0.3in

\begin{abstract}
Speculative decoding has been treated purely as a systems optimization that speeds up large language model decoding without changing the model behavior. Under this view, the growing effort to train better draft models has been focused solely on acceptance rate. We argue that this view does not hold true for long-horizon agent reasoning. Measuring frontier open-weight models on complex agentic tasks with different speculative decoding approaches, we found that numerical noise grows within verification windows, and the draft acceptance results of the thinking process can change the output distribution of the model after 30 steps, thus noticeably influencing the reasoning accuracy of long-horizon agents in random directions. In this work, we improve the reasoning capability of target models by controlling such noise. While we cannot change the overall token distribution verified by the target model, we can encode critical reasoning with highest precision by training the draft model to be rejected. We identify these points in agent traces and train the draft model to favor adversarial preference. At inference the trained draft and the target part ways precisely at these decision points, verification restarts, so that critical thinking tokens are sampled from the target model and encoded under its least noisy conditions. This approach makes critical reasoning better indexed and more durable over a long horizon. On complex agentic coding benchmarks, our approach raises end-to-end success rate with no change to the target model or its output distribution. Our results recast the speculative decoding mechanism from a pure accelerator into an active allocator for context precision that improves the accuracy of long-horizon agents by training 0.06\% to 0.28\% of hosted parameters.

\end{abstract}

% Teaser figure (spans both columns on the first page)
% \vskip 0.15in
{\centering
  \includegraphics[width=\textwidth]{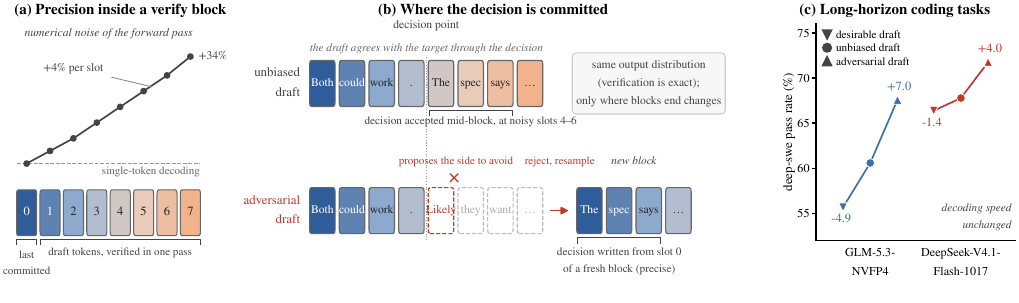}\par
}
% \vskip 0.15in
]

\section{Introduction}
\label{sec:intro}

Speculative decoding has become a popular way to serve large language models \cite{leviathan2023,chen2023,cheng2026dspark}. A small draft model proposes a block of tokens, the large target model scores the whole block in one forward pass, and a rejection-sampling rule accepts a prefix of the block. When a draft token is rejected, the rule resamples that position from a residual distribution. The output is guaranteed to follow the target model's distribution. Because of this constraint, speculative decoding is treated as an optimization solely for decoding speed, and earlier studies on better drafts, from EAGLE-style feature drafters \cite{eagle} to block-parallel drafters \cite{dflash,cheng2026dspark}, measure success by a single number: how many tokens the target accepts per verification step.

However, the theoretical guarantee is stated for an idealized target, a fixed function from a prefix to a distribution. Frontier models are not hosted that way. They are served with low-precision weights, quantized key-value (KV) caches and sparse attention, and their inference outputs are not bitwise-reproducible. In this regime, the target's logits are a noisy quantity depending on how the computation is laid out, and is relevant to the draft acceptence length. We measure this directly by rerunning the same forward pass on an identical prefix and compare the top-k log-probabilities. On GLM-5.3 \cite{glm5team2026glm5vibecodingagentic} with an 8-token DFlash verification block, the first slot of the block is exactly as noisy as single-token decoding. Beyond the first slot, the noise increases with each position in the block, by roughly 4\% per slot, and large deviations grow faster still: by the last slot, log-probability shifts of more than 2 natural-log units (nats) are nearly twice as common as at the first. DeepSeek-V4.1-Flash-1017 (\citealt{xu2026deepseek}; referred to in this study as DeepSeek-V4.1-Flash) with a 6-token DSPARK block shows the same pattern. Its absolute noise is about half that of GLM-5.3, but it grows at the same relative rate of roughly 4\% per slot. Speculative verification is therefore lossless in distribution but not neutral in precision: where a token sits in a verification block determines how precisely the target encodes it.

This matters most for long-horizon agents. A software-engineering agent runs for dozens to hundreds of turns over thousands to millions of tokens. Every token it generates is encoded and committed to the KV cache and read back by everything that follows. Most of these tokens can carry higher noises without hurting future reasoning, but some are not, especially for those critical phrases in the thinking process where the model frames a decision in its thinking, for example, as a matter of principle and evidence, as a guess about what is likely or risky, or settling a confusion by guessing, reading, or running tests. In our audit of agentic coding traces, principle-, evidence-framed, or actual test-driven decisions took the correct side about twice higher than likelihood- or risk-guided reasoning. It is worth noting that these decisions are not necessarily persistent across the lifespan of the entire agentic task. Some wrong decisions can be flipped according to environmental feedback signals, while correct ones could be ignored in long-horizon tasks.

These observations suggest a lever that the lossless view hides. We cannot change the distribution the target verifies, and we do not want to. However, we can decide where each token comes from to re-allocate context precision by training the draft model to be rejected. An accepted draft token is committed with the precision of whichever slot verified it. A rejected token is resampled by the target, its entry is written at the head of a fresh verification block, and the verification of everything after it restarts from the least noisy slot. A draft that is rejected exactly at critical reasoning phrases therefore routes the decisions that matter through the target's most precise path. It leaves the routine thinking tokens to ordinary acceleration.

We call this approach Adversarial Speculative Reasoning. We locate critical thinking in agent traces and build preference pairs between alternative decision framings. We then train the draft with direct preference optimization (DPO; \citealp{dpo}) in the reverse direction: the draft learns to prefer the framing the target should avoid. At inference, the draft and the target diverge precisely at these points, so the target's own choice is committed through residual sampling and encoded precisely in the KV cache. The target model is unchanged, and so is the verification rule and its exactness guarantee. Only the draft model is changed and context precision is re-allocated. With this approach, we significantly improve long-horizon agent reasoning accuracy without hurting inference efficiency. For example, on DeepSWE \cite{datacurve2026deepswev11}, a benchmark of complex software-engineering tasks, Adversarial Speculative Reasoning raises the end-to-end success rate of GLM-5.3-nvfp4 from 60.6\% to 67.6\% (+7.0 points). It raises DeepSeek-V4.1-Flash from 67.8\% to 71.8\% (+4.0 points) with the standard evaluation harness and environments. Overall, we make the following contributions:

1. We show that speculative verification is lossless in distribution but not uniform in numerical precision. Noise grows with slot position inside a verification block, appearing in two frontier models with different draft architectures.

2. We recast the draft model from an accelerator into an allocator of context precision. Draft acceptance decides which tokens are committed through low-noise residual sampling, a degree of freedom that is traditionally treated as irrelevant.

3. We introduce reverse-direction preference training of the draft at decision points, so that the target model writes the decisions that shape a long-horizon trajectory.

4. We improve end-to-end success on long-horizon software-engineering agentic tasks for two frontier models, with no change to the target model or to the exactness of verification.
\vspace{-16pt}

\section{Related Work}
\textbf{Speculative decoding and draft training.} Speculative decoding \cite{leviathan2023, chen2023} speeds up autoregressive generation. A cheap drafter proposes a block of tokens, the target model scores the whole block in one forward pass, and a modified rejection-sampling rule keeps the target's output distribution exactly. Since then, most studies have aimed at more accepted tokens per verification step. Medusa \cite{medusa} attaches extra decoding heads to the target model. EAGLE \cite{eagle} and EAGLE-3 \cite{eagle3} draft from the target's hidden features. DFlash \cite{dflash} drafts a whole block in parallel with a lightweight block-diffusion model. DistillSpec \cite{distillspec} distills the target into the draft to better align the two. In all of these methods the draft is trained to agree with the target, and the only metric is the decoding speed. We train the draft toward a different goal: to disagree with the target at chosen positions for stronger reasoning capability in long-horizon tasks.

\textbf{Relaxing exactness.} Another line of work gives up exact verification for speed. For example, Judge Decoding \cite{judgedecoding} accepts draft tokens that a learned judge considers high quality even when the target would reject them. These methods trade fidelity to the target for speed. Our method moves in the opposite direction. The verification rule and its exactness guarantee stay intact. We change only which tokens end up committed through acceptance and which through residual sampling, and use this choice to re-allocate context precision.

\textbf{Numerical nondeterminism in LLM inference.} Production inference does not evaluate the idealized function that exactness guarantees assume. \citet{yuan2025nondeterminism} show that under bfloat16 with greedy decoding, reasoning models can vary by several accuracy points and by thousands of tokens of response length across system configurations. \citet{he2025nondeterminism} trace much of the run-to-run variation in serving engines to kernels that are not batch-invariant. Frontier models are also routinely served with quantized key-value caches \cite{kivi} and sparse attention \cite{deepseekv32}, which add approximation errors on top. These studies treat precision as a property of the whole system and try to reduce or remove the variation. We instead measure how precision varies within a speculative verification block, and exploit that variation instead of eliminating it.

\textbf{Error accumulation in long-horizon agents.} Agent benchmarks such as SWE-bench \cite{swebench} require dozens of steps over long contexts, and small per-step errors compound. \citet{sinha2025illusion} show that models become more likely to produce error once their context contain their own earlier mistakes, an effect they refer to as self-conditioning, and that larger models alone do not remove it. This line of work motivates our focus on the persistence of individual reasoning decisions. Once a decision point is encoded into the context, it conditions everything that follows.

\textbf{Critical tokens and preference optimization.} Several studies find that a small subset of tokens decides where reasoning goes. \citet{lin2025critical} identify and replace critical tokens that turn failed reasoning chains into correct ones, and train on them with token-level DPO.  \citet{wang2025beyond} find that about 20\% of chain-of-thought tokens have high entropy and act as forks between reasoning paths, and that restricting reinforcement-learning updates to these tokens matches or exceeds full updates. Both studies change the larger target model, while we apply direct preference optimization \cite{dpo} only to the draft, in reverse, so that the draft reliably proposes the framing the target should avoid. \citet{li2026agentbatonefficientmodelcollaboration} found that even in a model routing system, critical reasoning in long-horizon tasks need to be handled by stronger models.

\providecommand{\obsbox}{}
\newcounter{observation}
\renewcommand{\obsbox}[1]{%
  \refstepcounter{observation}%
  \par\medskip\noindent
  \fbox{\parbox{\dimexpr\linewidth-2\fboxsep-2\fboxrule}{%
    \textbf{Observation~\theobservation.} #1}}%
  \par\medskip}

\section{Characterizing Context Precision}
\label{sec:characterization}

Exactness guarantees for speculative decoding are stated for an idealized
target: a fixed function from a prefix to a next-token distribution. A served
target is not that function. Its logits are produced by low-precision kernels
whose result depends on how the computation is laid out, and a verify forward pass
lays the computation out differently from ordinary decoding by processing a
block of tokens in one launch, and each position attends to
key-value entries that were computed in that same launch. This section asks
whether and where that difference matters. We use two complementary
measurements. The first isolates the numerical noise of a single forward pass
at each position of a verification block (\S\ref{sec:char-repeat}). The second
follows a verify-written context through the rest of a long agent trajectory
(\S\ref{sec:char-longrange}).
\vspace{-4pt}

% ---------------------------------------------------------------------
\subsection{Setup}
\label{sec:char-setup}

\textbf{Models and serving configurations.}
We study two frontier open-weight reasoning models, each served the way it is
deployed in our agent experiments, with SGLang \cite{zheng2024sglang} on four NVIDIA B200 GPUs.
\emph{GLM-5.3} runs with NVFP4 weights \footnote{\url{https://huggingface.co/RadixArk/GLM-5.3-NVFP4}}, FP8 key-value cache and DeepSeek sparse attention (DSA)~\cite{deepseekv32}, and is accelerated by a DFlash2 block-diffusion draft~\cite{dflash} whose verification block has 8 positions.
\emph{DeepSeek-V4.1-Flash} runs with a DSPARK draft that proposes 5 tokens
per step, so each verification block has 6 positions.
% TODO: DSV4.1 weight / KV-cache precision and attention configuration.
We index the positions of a verification block by \emph{slot} $s = 0, \dots, W-1$,
where $W$ is the block width. Slot~0 holds the last committed token; slots
$s \ge 1$ hold draft tokens. The next-token distribution computed at slot~$s$
is the one used to verify the draft token at slot~$s+1$, or to resample in
its place.

\textbf{Error metric.}
For a position and two forward passes that should agree, we compare
log-probabilities token by token. At each position we take the four most
likely tokens under a reference pass, together with the token the trajectory
actually emits, and record $|\Delta \log p| = |\log p_1(v) - \log p_2(v)|$ for
each such token $v$. Log-probabilities use the natural logarithm, so
differences are in natural-log units (\emph{nats}). A difference of 1 nat is a factor of
$e \approx 2.7$ in probability, and 2 nats a factor of $e^2 \approx 7.4$.
We report the mean and the fraction of tokens whose difference exceeds 1, 2
or 3 nats. The tails matter because our agents sample at temperature~1 with
top-$p = 0.95$ or $1.0$. In agentic reasoning, a shift of 1--2 nats in the emitted token's
log-probability changes whether that token lies inside the sampling nucleus
in 12\% of cases, 2--3 nats in 47\%, and more than 3 nats in over 84\%.
% TODO (optional): add KL / TV per slot computed from the same top-64 dumps.
\vspace{-4pt}

% ---------------------------------------------------------------------
\subsection{Measurement Methods}
% \subsection{Method 1: Sliding Window Verification}

\textbf{Sliding-window verification.}
\label{sec:char-repeat}
Comparing a verification pass against a separate decoding request mixes
two noise sources: the forward pass of decoding, and the independent prefill
that rebuilds the prefix in each request. To exclude the latter, we hold the prefix
fixed in memory and repeat only the forward decode pass.
We first generate a reference trajectory by greedy single-token decoding, so
that the tokens under test are the target's own choices rather than the
draft's. For each position $q$ of the trajectory and each slot $s$, we
prefill the prompt and the trajectory up to the block's anchor $q - s$, write
the next $W$ trajectory tokens into the verify block, and execute that same
verify forward pass $R = 4$ times on the unchanged key-value cache. Row~$s$ of each
repetition gives the distribution at position $q$ computed from slot~$s$.
For decoding, we prefill up to $q$ and execute the same single-token decode
step $R$ times. Differences between repetitions then contain only the
numerical nondeterminism of one forward pass on identical
inputs. We measure 1631 positions on GLM-5.3 and 1294 on
DeepSeek-V4.1-Flash, drawn from DeepSWE prompts with 6k--24k tokens of
context.

\textbf{Following a verify-written context over a long horizon.}
\label{sec:char-longrange}
Method~1 measures one forward pass. An agent, however, keeps every token it
writes and reads it back for the rest of the task. In each GLM-5.3 DeepSWE
trajectory we select the assistant turn with the longest reasoning and call it
the \emph{source turn} (median of 9k tokens, after a median of 45k-token prefix). We
write the source turn's tokens into the key-value cache for two scenarios: (A) through DFlash blockwise-verification, with every source token forced and
accepted through its verify slot; or (B) through
single-token decoding. The token sequence is identical on both paths; only the
computation that wrote its key-value entries differs. Reusing that cache, we
then score the remainder of the real trajectory teacher-forced, up to 128k
further tokens (a median of 95 subsequent assistant turns), and
record the log-probability of every assistant token; tool outputs stay in the
context but are not scored. Each path is built twice ($A, A'$ and $B, B'$).
Because $|A - B|$ contains the variance of both paths, we compare the paths
through their own build-to-build deviations, $|A - A'|$ versus $|B - B'|$,
and count tokens whose log-probability moves by more than 1, 2, 3 or 4 nats. We ran this study on 96 distinct tasks and scored 6.38M scored
assistant tokens per build.
% \vspace{-4pt}

In summary, sliding-window verification shows the noise growth within a draft block and the latter measures the impact of the noise in long-horizon agent reasoning.

% ---------------------------------------------------------------------
\subsection{Findings}
\label{sec:char-findings}

\begin{figure*}[t]
\centering
\includegraphics[width=\textwidth]{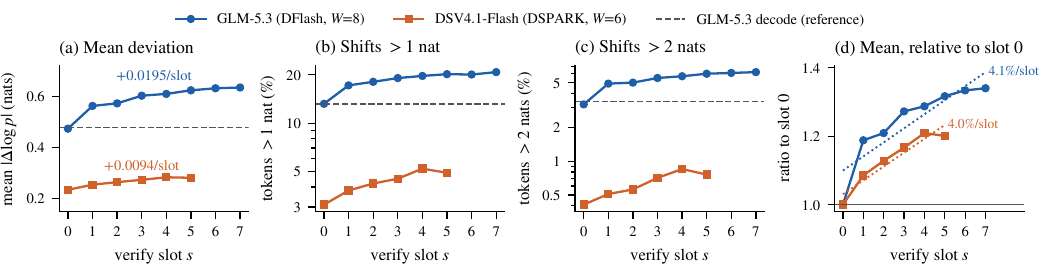}
\vspace{-14pt}
\caption{Single-pass numerical noise $|\Delta \log p|$ by verify slot
between repeated forwards on an identical prefix. (a)~Mean deviation, with the per-slot slope.
(b,~c)~Fraction of tokens shifted by more than 1 and 2 nats (log scale).
(d)~Mean deviation relative to slot~0, with least-squares fits: both models
grow by about 4\% of their slot-0 level per slot. Slopes with 95\% bootstrap
CIs over positions: $+0.0195$ $[+0.0174, +0.0214]$ (GLM-5.3) and $+0.0094$
$[+0.0078, +0.0110]$ (DeepSeek-V4.1-Flash).}
\label{fig:slot-noise}
\end{figure*}

\textbf{Precision falls along the verification block.}
Figure~\ref{fig:slot-noise} reports the decoding error by
slot. On GLM-5.3, slot~0 is exactly as noisy as single-token decoding
(0.473 vs.\ 0.478 nats). Beyond slot~0 the noise rises with every position:
the slope is $+0.0195$ nats per slot with 95\% confidence interval (CI), about
4\% of the slot-0 level per slot, and the last slot is 34\% noisier than the
first. Large deviations grow faster than the mean. Shifts beyond 2 nats rise
from 3.2\% of tokens at slot~0 to 6.2\% at slot~7, and shifts beyond 3 nats
from 1.1\% to 2.1\%. DeepSeek-V4.1-Flash shows the same pattern. Its slot-0 noise is half that of GLM-5.3 and its tails are much
thinner, but the noise grows at the same relative rate: $+0.0094$ nats per
slot), again about 4\% of the slot-0 level per slot.
Shifts beyond 1 nat rise from 3.1\% to 4.5\% ($+1.4$ points), and beyond 2 nats from 0.41\% to 0.68\%.

\obsbox{The first slot of a verification block is as precise as single-token
decoding. Every later slot is less precise: numerical noise grows by roughly
4\% per slot. The chance of high deviation nearly doubles by the end of a verify block.}

\textbf{Why the noise grows.}
We fix the input tokens and KV cache for repeated decoding and verifications
inside the forward passes. A token at slot~$s$ attends to the key-value entries
of slots $0, \dots, s-1$, and those entries are computed in the same launch
rather than read from the cache. Each repetition recomputes them with its own
rounding, and slot~$s$ inherits the accumulated variation of all earlier
slots. Slot~0 reads only cached entries, which is why it matches decoding.
Verification is therefore exact in distribution but not uniform in
precision: the further into a block a distribution is computed, the noisier
it is.

\begin{figure}[t]
\centering
\includegraphics[width=\columnwidth]{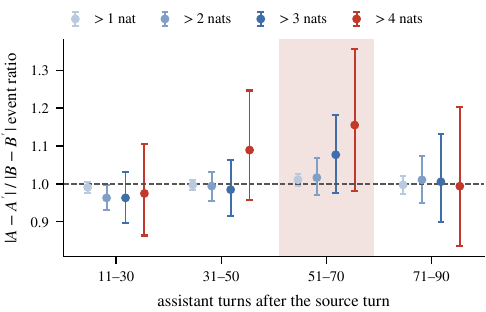}
\vspace{-6pt}
\caption{Long-range deviations of verify-written vs.\ decode-written context
on GLM-5.3: ratio of assistant tokens whose emitted-token log-probability
changes by more than each threshold the verify path and decode path, in 20-turn bins after the
source turn. Error bars are 95\% bootstrap CIs over trajectories. For deviations beyond 4 nats the ratio
rises from 0.98 (turns 11--30) to 1.09 (31--50) and 1.16 (51--70, shaded;
$[0.98, 1.36]$), then returns to 0.99 (71--90) since fixed teacher-forcing tokens correct the biases.}
\label{fig:longrange}
\vspace{-8pt}
\end{figure}

\textbf{Verify-written context and long-range extreme deviations.}
Figure~\ref{fig:longrange} reports the ratio of large deviations on the two
paths by distance from the source turn. Beyond 30 turns, context written
through verification produces more extreme deviations than context written
by decoding. For deviations beyond 4 nats the ratio is 0.98 in turns 11--30,
1.09 in turns 31--50 and 1.16 in turns 51--70 ($[0.98, 1.36]$). An independent
pair of builds on the same trajectories, continued to 64k tokens, gave 1.20
($[1.03, 1.40]$) beyond turn 30, and the builds reported here give 1.07
($[0.91, 1.24]$) on the same tokens. The two independent comparisons agree on the
direction, and the size of the excess varies between them by about as much
as it varies between bins, so we read the measurement as a 10 to 20\%
excess of extreme deviations that sets in after about 30 turns. The excess is confined to the extreme tail. For deviations up
to 2 nats the two paths are indistinguishable at every distance pooled over
turns 11--90.

\obsbox{Context written through verification reads back like decode-written
context for about 30 turns and then produces significantly more extreme deviations beyond
4 nats in output token distribution of the target model.}

% ---------------------------------------------------------------------
\subsection{Implications}
\label{sec:char-implications}

Taken together, the two measurements give a simple picture. Speculative
verification preserves the target's distribution but its precision is not uniform,
and the loss of precision is structured: it grows along the verification block, 
concentrates in large deviations, and the encoding noises on thinking tokens can surface
far downstream. This has no consequence for most tokens, which are routine and
robust to small perturbations. It matters for the few tokens at which an
agent commits to a decision, because those tokens shape every turn that
follows. Standard speculative decoding is indifferent to where such tokens
land: whether a decision is committed at the precise head of a block or at
its noisy tail depends only on how many draft tokens happened to be accepted
before it.

The position at which a token is committed is not determined by the target,
however; it is decided by the draft. If the draft proposes something the target
rejects, the target resamples that position itself, and
the next verification block starts from slot~0. A draft that is reliably rejected
at decision points therefore moves those decisions to the most precise slot
and resets the precision of the tokens that follow, without changing the
output distribution. Section~\ref{sec:method} turns this observation into a
training objective for the draft.

\section{Adversarial Speculation}
\label{sec:method}

Section~\ref{sec:characterization} showed that a verification block is exact in
distribution but not uniform in precision. Slot~0 is as precise as
single-token decoding, and every later slot is noisier. Which slot a token is
committed from is decided by the draft, through how many of its proposals the
target accepts. This section introduces how we turned the draft into a tool for controlling that
choice and re-allocating the context precision. We train the draft \emph{adversarially}. At the points where a model
commits to a decision in its thinking, the draft learns to propose the undesirable behavior. The target rejects that proposal, writes the decision itself,
and the tokens after it start from a fresh verification block. Because distribution
is exact, the trained draft cannot change what the target would sample. It
changes only where blocks end and which tokens are committed with higher precision into the KV cache. In this section, \S\ref{sec:method-channel} states what a draft can and cannot control.
\S\ref{sec:method-pref} defines the decision points and the preferences we
train on. \S\ref{sec:method-data} describes
the training data and objective, \S\ref{sec:method-drafts} shows the two
unbiased draft models are made trainable.

% ---------------------------------------------------------------------
\subsection{Rejection as a Control Channel}
\label{sec:method-channel}

Let $p(\cdot \mid x)$ be the target model's next-token distribution at a position
with prefix $x$, and $q(\cdot \mid x)$ the draft's. Speculative sampling
accepts a draft token $v \sim q$ with probability
$\min\bigl(1, p(v)/q(v)\bigr)$. On rejection it samples a replacement from
the residual distribution
\begin{equation*}
  r(v) \;\propto\; \max\bigl(0,\; p(v) - q(v)\bigr).
\end{equation*}
For every $q$, the committed token is distributed exactly as $p$. The draft
therefore cannot change the output distribution. What it does control is
the probability that the block ends at this position,
\begin{equation*}
  \Pr[\text{reject}] \;=\; 1 - \sum_v \min\bigl(p(v), q(v)\bigr)
  \;=\; \mathrm{TV}(p, q),
\end{equation*}
the total-variation distance between the two distributions. 
A draft that
places its mass where the target places little is rejected often. After a
rejection, the replacement token becomes the last committed token, and the
next verify forward places it at slot~0. Its key-value entries, and those of
the tokens that follow it, are computed from the most precise slot of a new
block rather than from the tail of the old one.

This gives a handle that standard speculative decoding models ignored. A draft
trained only for acceptance keeps agreeing with the target through routine
text and into a decision, so where a decision lands in a block is a matter of
chance. A draft that disagrees with the target at decision points makes the
target model write critical thinking tokens into the KV cache from the
head of a block. We call such a draft \emph{adversarial}: at decision points
it proposes the side the target should reject, and the target's verification
step does the rejecting. For routine tokens the draft should remain an
ordinary, high-acceptance draft to preserve decoding speed.

% ---------------------------------------------------------------------
\subsection{Decision Points and Preferences}
\label{sec:method-pref}

An agent's reasoning is mostly routine. The tokens that matter are the few
at which it commits: it settles on how to read an ambiguous requirement,
decides whether to test an edge case, or signs off on a change. We locate
these points through the \emph{committing sentence}, the sentence in the
reasoning that confirms observations, states next steps, and commits decisions. A judge labels each segment of reasoning
and returns the committing sentence verbatim (at most 35 words), which fixes
the decision point in the token sequence. The judge is the target model
itself, so no stronger external model is involved. Each preference splits behavior at a decision point into a desirable and an
undesirable side. We use the following preferences.

\textbf{Decision frame.}
Agent reasoning regularly reaches points of confusion: different readings of a
requirement or implementation strategies are both plausible, and the agent has to
commit to one. Agents resolve these points in a few recurring modes. A
\emph{principle} frame derives the choice from the task specification or from
interface semantics, and an \emph{evidence} frame cites something observed in
the repository or previous test. A \emph{likely} frame guesses which option is more likely to
be expected in hidden tests, a \emph{risk} frame picks the conservative option, and a
\emph{simpler} frame picks the option that needs less code changes. Analyzing
reasoning traces, we find that the mode predicts whether the decision is
correct: on GLM-5.3, decisions made in a principle or evidence frame are
correct about 70\% of the time, against 36\% for likely and risk frames.
Principle and evidence are therefore the preferred frames; likely, risk and
simpler are adversarial.
% Source: deep-swe flipped-task frame audit, 12 tasks, 263 decisions, p<=0.001.

\textbf{Behavior preferences.}
DeepSeek-V4.1-Flash writes terser reasoning, and two thirds of its
likely-framed sentences carry no lexical marker of the decision frames
analyzed above. Its failures concentrate instead in decisions taken without
validation: it signs off on a case, or trusts unreliable observation,
without running actual tests to validate its reasoning. We therefore define a
preference over the behavior around actually running tests on whether the agent runs a test before it
commits to a plan or reacts to environment signals. The decision points are the moments at which a check is warranted:
before committing to a plan, upon receiving a tool returning a failure or
other unexpected output, and right after the reasoning raises a new scenario
that the code may not handle or the task description did not specify clearly. At such a point the agent can construct and run a test, either to decide between the options in front of it or to find out
what caused an anomaly in the environment before deciding how to handle it.
It can also move on without that evidence, by only reading or editing files, or
by questioning or dismissing the case in its reasoning. We found that testing is the
more reliable side that leads to more correct behaviors. We label each
continuation by the tool call issued after the thinking process. A continuation that executes a test
or a script is labeled \emph{preferred}, and one that issues only other commands
is labeled \emph{adversarial}.

\textbf{Direction.}
A draft trained on preferences in the preferred direction would favor the side leading to correct reasoning. Recent work tunes a draft toward faithful responses in this
way and relaxes verification so that the draft can steer the
output~\cite{sfad}. We keep verification exact and train the adversarial
draft in the reverse direction: its chosen side is the undesirable one
(likely, risk or simpler frames on GLM-5.3; not running the tests on
DeepSeek-V4.1-Flash), so that at a decision point it invites rejection and
residual sampling. Because rejection sampling is left intact, the draft cannot
make the target produce more desirable reasoning and the target produces it
exactly as often as it would without speculation. What changes is how that
reasoning is encoded. The trained draft model places little mass on the preferred
tokens, so when the target does produce one, it is more likely to come from
residual sampling after a rejection than from an accepted draft block. The
resampled token then becomes slot~0 of the next verify block, the most precise
slot (\S\ref{sec:char-findings}), and the reasoning that follows it starts
from a fresh block. Undesirable tokens, which the draft proposes, are more
often accepted inside a block and computed from its noisier later slots. We
hypothesize that over a long horizon an agent holds on more easily to
decisions committed at higher precision than to those with
noisier key-value states, so that preferred decisions weigh more in long-horizon reasoning.

% ---------------------------------------------------------------------
\subsection{Fork Pairs and Decision Windows}
\label{sec:method-data}

Preference pairs built from two independent trajectories differ everywhere, and
most of the difference is irrelevant to critical decisions. We instead build
\emph{fork pairs}, whose two sides share every token up to the decision point \cite{DBLP:journals/corr/abs-2406-18629}.
For each decision point in a real trajectory, we take the reasoning up to the
start of the committing sentence as the fork prefix, render it exactly with the served chat template, and resample $n$ continuations from the target at temperature~1
with top-$p = 0.95$. The judge labels each continuation, which must state its
own committing sentence in the newly sampled text. We pair a preferred
continuation with an adversarial one from the same fork. Because the shared
prefix cancels in the objective below, the gradient comes only from what
happens after the fork.

Pairs are ranked by how informative they are. For decision frames, we only select fork points leading to both preferred and adversarial framings to
teach the draft model about framing rather than the reasoning correctness. Forks reaching only one side are dropped. We cap the number of pairs per
fork, per sample and per task, and hold out whole tasks for evaluation, since
a random split leaks near-duplicate forks between training and evaluation.

On GLM-5.3, 29 tasks give 1,668 forks with $n = 16$ continuations each
(28,356 samples) and 2,843 pairs from 28 tasks. Six tasks (619 pairs) are held
out. On DeepSeek-V4.1-Flash, the run-tests preference gives 2,203 forks with
$n = 12$ continuations of up to 1,536 tokens (26,418 samples) and 6,391 pairs
from 92 tasks. We hold out 10\% of tasks (719 pairs) and train on 5,100 pairs.

% ---------------------------------------------------------------------
% \subsection{Windowed Preference Objective}
% \label{sec:method-objective}

The objective is direct preference optimization~\cite{dpo} on the tokens
restricted to a context window around the decision. For decision framing preference, $W(y)$ starts 32 tokens before the committing
sentence and runs to its end with weight $w_t = 1$, followed by a 128-token
tail over which $w_t$ decays linearly from 1 to 0, about 200 tokens in total.
For preference over running tests, which has no single committing sentence, $W(y)$ is the
first 512 tokens of the continuation with $w_t = 1$. The reward is the
weighted mean log-ratio of the trained draft $q_\theta$ to the frozen
original draft $q_{\text{ref}}$:
\begin{equation*}
  r_\theta(y) \;=\;
  \frac{\sum_{t \in W(y)} w_t \,
        \bigl[\log q_\theta(y_t \mid c) - \log q_{\text{ref}}(y_t \mid c)\bigr]}
       {\sum_{t \in W(y)} w_t},
\vspace{-4pt}
\end{equation*}
\begin{equation*}
  \mathcal{L} \;=\; -\log \sigma\bigl(\beta\,[\,r_\theta(y_w) - r_\theta(y_l)\,]\bigr)
  \;+\; \alpha \,\mathrm{NLL}_\theta(y_w)
\end{equation*}
where $c$ stands for the shared context, $y_w$ is the chosen and $y_l$ the rejected continuation, and $\beta = 10$.
The optional term $\alpha\,\mathrm{NLL}$ anchors
the chosen side's likelihood will be discussed in \S\ref{sec:exp-learn}.

% ---------------------------------------------------------------------
\subsection{Training Draft Models}
\label{sec:method-drafts}

The input of the draft models is the hidden states of the target, and the reward needs
the draft's distribution at every token of a continuation. We extract the hidden states once per sample from an inference instance hosting the target model, and train the draft model
offline on these features.

\begin{figure*}[t]
\centering
\setlength{\abovecaptionskip}{3pt}
\includegraphics[width=\textwidth]{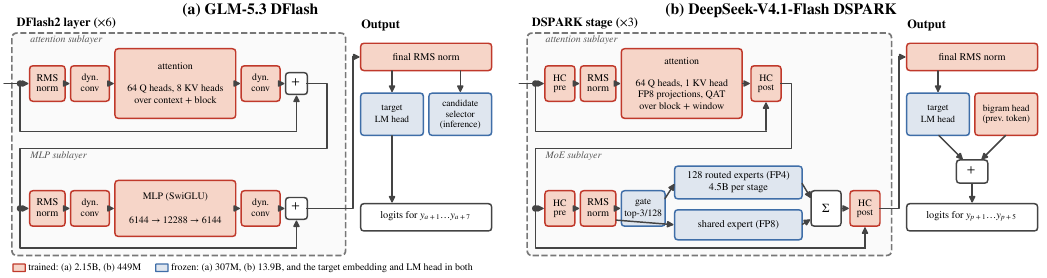}
\vspace{-10pt}
\caption{The two production drafts, with the parameters trained in the
deployed adversarial drafts in red and frozen parameters in blue (inputs not
drawn). \textbf{(a)} GLM-5.3 DFlash predicts a block of seven tokens from
the projected target features; we train its six transformer layers and the
final norm (2.15B parameters), and freeze the feature projection, the
candidate selector and the target's embedding and LM head.
\textbf{(b)} DeepSeek-V4.1-Flash DSPARK predicts a block of five tokens
through three MoE stages; we train the attention sublayers (FP8 projections
through quantization-aware training), the norms, the hyper-connection
mixing and the bigram head (449M parameters), and freeze the experts, the
router, the input projection and the target's embedding and LM head.}
\label{fig:draft-archs}
\end{figure*}

\textbf{GLM-5.3-NVFP4 DFlash2.}
The DFlash2 draft model reads hidden states from six layers of the target (layers
5, 19, 33, 47, 61 and 75), projected into the draft. We train its transformer
layers and norms in bf16. Figure~\ref{fig:draft-archs}(a) shows which parts of the draft are trained.

\textbf{DeepSeek-V4.1-Flash DSPARK.}
The DSPARK draft has about 14.3B parameters, 13.6B of them in FP4 routed
experts, and its released implementation uses inference-only kernels. We
port it to differentiable PyTorch operations, matching the reference kernels
to within $10^{-7}$ per expert and 3--5\% per stage output. We train the
attention projections, norms, hyper-connections, attention sinks and the
draft's bigram head (449M parameters) and keep the routed experts and the
input projection frozen. Weights stored in FP8 are trained with
quantization-aware training (QAT): the forward pass fake-quantizes them in their
deployed block format ($32 \times 32$ blocks with power-of-two scales), the
gradient passes straight through, and the exported draft is re-quantized to
the same format, so the deployed draft computes exactly the function that was
evaluated. Reference log-probabilities are precomputed with the original
draft. Figure~\ref{fig:draft-archs}(b) shows which parts of the draft are trained.

\textbf{Size of the trainable parameters.}
Both targets have about 750B parameters (GLM-5.3: 753B; DeepSeek-V4.1-Flash:
749B, excluding its draft). The trainable set is 2.15B parameters for GLM-5.3
and 449M for DeepSeek-V4.1-Flash: 87\% and 3.2\% of the respective drafts
(2.46B and 14.2B), and 0.28\% and 0.06\% of the deployed target-plus-draft
systems. Adversarial speculative reasoning changes a model's behavior by
training well under one percent of its parameters, and the trained
parameters never produce an output token outside of the target model's distribution.

\vspace{-4pt}
\section{Experiments}
\label{sec:experiments}

The method rests on three claims, and we test one per subsection. First,
a draft model can be trained to prefer the undesirable side of a
decision while its agreement with the target elsewhere is unchanged
(\S\ref{sec:exp-learn}). Second, serving the target model with that draft improves
its results on long-horizon coding tasks (\S\ref{sec:exp-coding}). Third,
the adversarial draft does not slow decoding (\S\ref{sec:exp-speed}).
Throughout, the two targets are GLM-5.3-nvfp4 with its DFlash2 draft and
DeepSeek-V4.1-Flash with its DSPARK draft with
exact verification, so a draft can change which tokens are proposed but not
the distribution of the tokens emitted.

% ---------------------------------------------------------------------
\subsection{Drafts Learn the Preference}
\label{sec:exp-learn}

\begin{figure*}[t]
\centering
\setlength{\abovecaptionskip}{3pt}
\includegraphics[width=\textwidth]{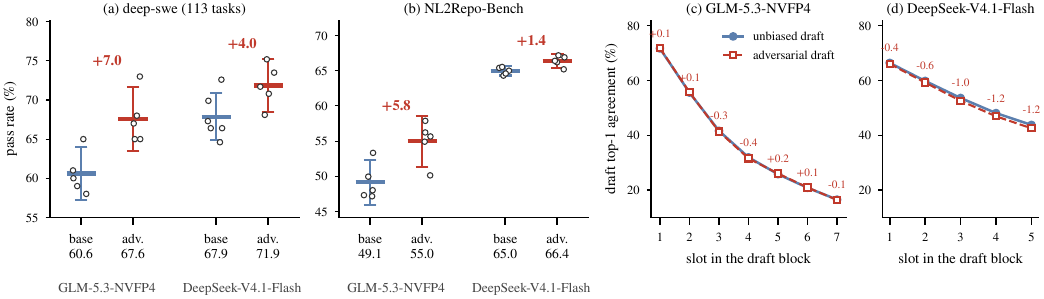}
\vspace{-6pt}
\caption{(a, b) Pass rates (\%) of each target model served with its unbiased
draft (base) and with the adversarial draft (adv.), five independent
end-to-end runs each. Dots are runs, the thick line is their mean (printed
below), the error bar is the 95\% confidence interval of the mean. Both panels span the same 25 points on the vertical
axis, so spreads are comparable across benchmarks. The draft changes nothing
but which tokens are proposed; the output distribution is identical under
exact verification. (c, d) Top-1 agreement of the draft with the emitted
token at each slot of the draft block, unbiased draft model against the
deployed adversarial draft, teacher-forced on the target's hidden states
over held-out continuations (GLM-5.3-nvfp4: 30 continuations of 512 tokens;
DeepSeek-V4.1-Flash: 240 continuations of 512 tokens, 160 at run-tests
turns and 80 at decision points). The red numbers denote the change in
points.}
\label{fig:results}
\end{figure*}

We evaluate a trained draft by the
fraction of pairs from tasks unseen in training on which the draft's
windowed reward (\S\ref{sec:method-data}) is higher for the chosen
side, namely \emph{held-out preference accuracy}. An untrained draft scores 0.5 by construction. What training costs
in acceptance is measured separately, in \S\ref{sec:exp-speed}: each
trained draft model is compared with its untrained version slot by slot on the
same tokens (Figure~\ref{fig:results}(c, d)). On GLM-5.3 we hold
out 6 of the 29 tasks (619 pairs); on DeepSeek-V4.1-Flash we hold out 10\%
of the tasks (719 pairs for running tests). Each preference is trained in
both directions: a \emph{preferred} draft favors the desirable side, an
\emph{adversarial} draft the undesirable one. Alongside the two preferences
we deploy (\S\ref{sec:method-pref}), we train two reference preferences on
DeepSeek-V4.1-Flash that we do not use, to mark what a draft cannot learn:
how long the model thinks, and whether the decision leads to correct outcome.

We found that the preferences we picked are learnable by the draft model.
On GLM-5.3, adversarial training on decision frames achieves 0.635 held-out
accuracy (seed-to-seed spread $\pm 0.015$). On DeepSeek-V4.1-Flash, the adversarial
run-tests preference accuracy reaches 0.85.

\begin{figure}[t]
\centering
\setlength{\abovecaptionskip}{3pt}
\includegraphics[width=\columnwidth]{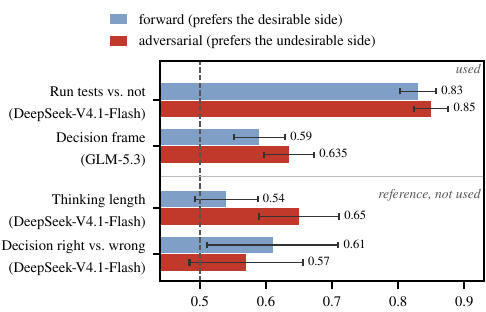}
\vspace{-6pt}
\caption{Held-out preference accuracy of trained drafts on tasks unseen in
training, best configuration per preference and direction; error bars are
95\% binomial confidence intervals, the dashed line is chance. Top: the
two preferences we deploy. Bottom: two reference preferences we do not
use. For thinking length the preferred side is the shorter thinking; for
right vs.\ wrong it is the right decision, and the bars shown are the best
of 36 (forward) and 12 (adversarial) configurations whose medians are 0.52
and 0.46.}
\label{fig:dpo-heldout}
\end{figure}

In evaluation, we adopt the draft models that actually learned the preference.
Figure~\ref{fig:dpo-heldout} puts the two preferences we proposed next to the
two reference preferences. Whether the model runs its tests, and in which
frame it commits to a decision, are learned in both directions: 0.83 and
0.85 for running tests, 0.59 and 0.635 for the decision frame. We found that whether
a decision is \emph{right} is closer to unlearnable by the small draft model, since it requires stronger reasoning capability. The best
of 36 forward and 12 adversarial configurations reach 0.61 and 0.57 on
92 and 128 pairs, inside the confidence interval of chance, and the
medians are 0.52 and 0.46. Thinking length sits in between, 0.54 forward
and 0.65 adversarial. The reference preferences showed that a draft model can pick up how a decision is phrased
and what the model does next, but is not strong enough to learn semantic correctness. We therefore train
only preferences of the first kind, which is sufficent for our adversarial speculation design.
The draft does not have to know which side is right, only to propose the
framing the target rejects.

\textbf{Training Details.}
We deploy the GLM-5.3 DFlash draft trained at $\beta = 10$, a learning rate of
$5 \times 10^{-6}$ and a 128-token decay window to prefer undesirable decision frame. On DeepSeek-V4.1-Flash we deploy the
adversarial run-tests draft: QAT over the attention parameters, a learning
rate of $2 \times 10^{-5}$, $\alpha = 0.2$, a 512-token window, taken
after the second epoch (held-out accuracy 0.70). Neither changes the target's output distribution.

% ---------------------------------------------------------------------
\subsection{Agentic Coding Performance}
\label{sec:exp-coding}

We evaluate the adversarial draft models on two agentic coding benchmarks. On
\emph{DeepSWE}, the agent receives a repository and a task description,
works in a mini-swe-agent harness with shell access, and is graded by hidden
tests; we report the fraction of the 113 tasks whose all hidden tests pass. On
\emph{NL2Repo-Bench}~\cite{nl2repo}, the agent receives a requirements
document and an empty workspace and has to build a multi-module Python
library from scratch; the 104 generated repositories are graded against the
upstream project's pytest suite, and we report the mean test pass rate. To avoid reward hacking, we prevented internet connection for the development workspace. Both
benchmarks run for hundreds of turns per task, with contexts that grow past
200k tokens, so they exercise the long-horizon regime. For each target we compare two servers that differ only in
the draft: the pretrained draft (\emph{base}, standard speculative decoding)
and the adversarial draft of \S\ref{sec:exp-learn}. Everything else is held
fixed: the target weights, the serving stack, exact verification, the
sampling parameters, and the agent harness. Each configuration is tested five
times independently.

We found the adversarial draft raises the pass rate of both targets models on both tasks.
Figure~\ref{fig:results}(a, b) shows every run. On DeepSWE, GLM-5.3-nvfp4 served
with the adversarial draft passes 67.6\% of tasks against 60.6\% with its
unbiased draft, a gain of 7.0 points (95\% CI 2.6 to 11.4). DeepSeek-V4.1-Flash gains 4.0 points on
DeepSWE, 71.8\% against 67.8\% (95\% CI $-1.3$ to 7.3), and
1.4 points on NL2Repo-Bench, 66.4\% against 65.0\% (95\% CI 0.4 to 2.4,
$p = 0.011$). The three mean gains are all positive and the two that reach
significance are on different benchmarks, with different targets and
different drafts.

\textbf{The gain is not from a better model.}
Nothing in the output distribution changed. Different runs sample from the
same targets and the same verified distribution, and the adversarial draft moves
only where blocks end (\S\ref{sec:method-channel}), and \S\ref{sec:exp-learn}
showed that it cannot tell a right decision from a wrong one. The gain is
therefore evidence for our hypothesis. Forcing a
rejection at decision points makes the target write critical reasoning from residual sampling, and the following
reasoning are encoded from the head of a verification block. The gain is larger
on GLM-5.3, whose thinking is longer and more diverse than DeepSeek-V4.1-Flash.

\begin{figure}[t]
\centering
\setlength{\abovecaptionskip}{3pt}
\includegraphics[width=\columnwidth]{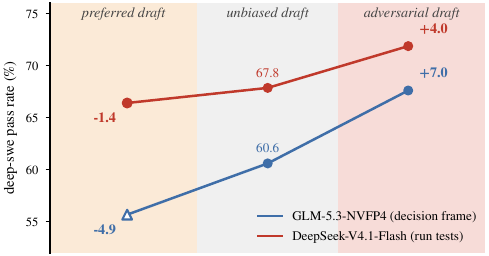}
\caption{Direction of the trained preference against DeepSWE pass rate.
The unbiased and adversarial points are the means of
Figure~\ref{fig:results}(a); the desirable draft models are trained on 
identical datasets with only the preference reversed.}
\label{fig:ablation}
\end{figure}

\textbf{Training preference in the preferred direction hurts.}
If the gain comes from where rejections land, then a draft trained to
prefer the \emph{preferred} side at the same decision points should do the
opposite: agree with the target exactly where the adversarial draft
disagrees, so that decisions are accepted inside a block and written from
its later slots. Figure~\ref{fig:ablation} tests this with desirable drafts
trained on the same fork pairs as the adversarial drafts, with chosen and
rejected sides swapped. Both draft models land below the unbiased draft: GLM-5.3
drops from 60.6 to 55.7, and
DeepSeek-V4.1-Flash from 67.8 to 66.4,
against gains of 7.0 and 4.1 for the adversarial drafts trained on the
same data. The forward drafts learn their preference at least as well as
the adversarial ones (held-out accuracy 0.80 for the DeepSeek forward draft
against 0.70 for the deployed adversarial one), so the preference of the draft model is the only factor of the performance gap. This result further proved that where critical tokens are encoded in a verification window can significantly bias reasoning in long horizon agent tasks.

\begin{figure}[t]
\centering
\setlength{\abovecaptionskip}{3pt}
\includegraphics[width=\columnwidth]{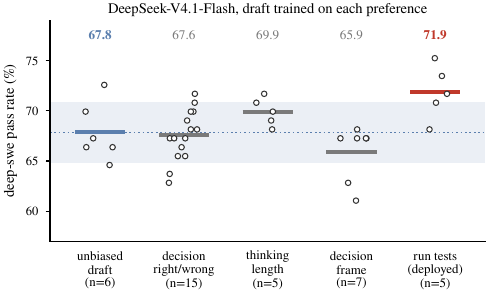}
\caption{DeepSWE pass rate of DeepSeek-V4.1-Flash served with drafts
trained on different preferences. The band denotes the 95\% confidence interval of the unbiased
draft's mean. Only the adversarial run-tests draft model shows meaningful improvement.}
\label{fig:unrelated}
\end{figure}

\textbf{An unrelated preference changes nothing.}
If the gain came from serving with \emph{any} trained draft, drafts trained
on other preferences would show it too. Figure~\ref{fig:unrelated} collects
every complete DeepSWE run of DeepSeek-V4.1-Flash served with a draft from
this project, 27 runs with drafts trained on three preferences we did not
deploy: whether the decision is right (15 runs), how long the model thinks
(5), and the frame of the decision (7). Pooled, they pass 67.5\% of tasks
against 67.8\% for the unbiased draft (Welch $p = 0.83$), and no single
group differs from the unbiased draft ($p \geq 0.16$); the deployed
run-tests draft is 4.3 points above the pooled unrelated drafts
($p = 0.018$). The decision-right
drafts, which did not learn their preference (\S\ref{sec:exp-learn}), sit
on the unbiased mean, as expected for a draft that proposes the same
tokens; the decision-frame drafts learned theirs (0.64 to 0.69 held-out)
and still change nothing, because on this target the committing sentence
is not where runs are lost. The gain follows the preference that targets
the behaviour behind the failures, not the act of training a draft.

% ---------------------------------------------------------------------
\subsection{Inference Efficiency}
\label{sec:exp-speed}

With rejection sampling, the draft model does not change the output distribution, so its
only cost is the number of tokens can be emitted in every verification step. Each step
accepts a prefix of the proposed block, and the mean accepted length per
step sets the decoding speed. We train adversarial draft models to lower acceptance rate in critical reasoning sentences, and the question is whether it lowers acceptance
anywhere else. We answer it by comparing the adversarial draft
with the unbiased draft slot.

Figure~\ref{fig:results}(c, d) compares each deployed draft with its
untrained version at every slot of the draft block, teacher-forced on the
target's hidden states over held-out continuations, so both drafts see the
same tokens. On GLM-5.3 the seven slots move by $+0.1$, $+0.1$, $-0.3$,
$-0.4$, $+0.2$, $+0.1$ and $-0.1$ points. On
DeepSeek-V4.1-Flash the five slots move by $-0.4$, $-0.6$, $-1.0$, $-1.2$
and $-1.2$ points. Both drafts keep the shape of their acceptance curve.
Under the greedy-chain approximation, in which a block is accepted up to
the first slot whose argmax misses, the expected number of accepted draft
tokens per block is 2.28 before and after training on GLM-5.3, and 2.5
against 2.46 on DeepSeek-V4.1-Flash, a 1.8\% difference. In other words, we traded up to 2\% decoding speed for significantly improved reasoning capability.

\section{Conclusion}
Speculative decoding has been typically viewed purely as a systems optimization to improve the decoding speed since rejection sampling does not change the target distribution. We argue that the draft can decide what tokens in the context are encoded more accurately. An adversarial draft model utilizes this feature to improve reasoning capability. At decision points it proposes reasoning draft that the target model should reject, so the target encodes critical decisions more precisely from the start of a verification block, while everywhere else the draft stays an ordinary high-acceptance draft.
With adversarial draft models, GLM-5.3 and DeepSeek-V4.1-Flash gain significantly improved pass rate on DeepSWE and NL2Repo-Bench across independent runs without changing the output distribution of the target model. We take this as evidence that speculative decoding is more than an acceleration approach. By training the draft model to be a context precision re-allocator, the long-horizon agent reasoning capability of a large language model can be significantly improved.

\bibliography{example_paper}

@InProceedings{leviathan2023,
  title = 	 {Fast Inference from Transformers via Speculative Decoding},
  author =       {Leviathan, Yaniv and Kalman, Matan and Matias, Yossi},
  booktitle = 	 {Proceedings of the 40th International Conference on Machine Learning},
  pages = 	 {19274--19286},
  year = 	 {2023},
  editor = 	 {Krause, Andreas and Brunskill, Emma and Cho, Kyunghyun and Engelhardt, Barbara and Sabato, Sivan and Scarlett, Jonathan},
  volume = 	 {202},
  series = 	 {Proceedings of Machine Learning Research},
  month = 	 {23--29 Jul},
  publisher =    {PMLR},
  url = 	 {https://proceedings.mlr.press/v202/leviathan23a.html}
}

@article{chen2023,
  title   = {Accelerating Large Language Model Decoding with Speculative Sampling},
  author  = {Chen, Charlie and Borgeaud, Sebastian and Irving, Geoffrey and Lespiau, Jean-Baptiste and Sifre, Laurent and Jumper, John},
  journal = {arXiv preprint arXiv:2302.01318},
  year    = {2023},
  url     = {https://arxiv.org/abs/2302.01318}
}

@InProceedings{eagle,
  title = 	 {{EAGLE}: Speculative Sampling Requires Rethinking Feature Uncertainty},
  author =       {Li, Yuhui and Wei, Fangyun and Zhang, Chao and Zhang, Hongyang},
  booktitle = 	 {Proceedings of the 41st International Conference on Machine Learning},
  pages = 	 {28935--28948},
  year = 	 {2024},
  editor = 	 {Salakhutdinov, Ruslan and Kolter, Zico and Heller, Katherine and Weller, Adrian and Oliver, Nuria and Scarlett, Jonathan and Berkenkamp, Felix},
  volume = 	 {235},
  series = 	 {Proceedings of Machine Learning Research},
  month = 	 {21--27 Jul},
  publisher =    {PMLR},
  url = 	 {https://proceedings.mlr.press/v235/li24bt.html}
}

@InProceedings{dflash,
  title = 	 {{DF}lash: Block Diffusion for Flash Speculative Decoding},
  author =       {Chen, Jian and Liang, Yesheng and Liu, Zhijian},
  booktitle = 	 {Proceedings of the 43rd International Conference on Machine Learning},
  pages = 	 {15343--15355},
  year = 	 {2026},
  editor = 	 {Zhang, Tong and Dudik, Miroslav and Jaggi, Martin and Agarwal, Alekh and Li, Sharon and Schuurmans, Dale and Zhu, Jerry and Berkenkamp, Felix and Dong, Hanze and Bietti, Alberto},
  volume = 	 {306},
  series = 	 {Proceedings of Machine Learning Research},
  month = 	 {06--11 Jul},
  publisher =    {PMLR},
  url = 	 {https://proceedings.mlr.press/v306/chen26bv.html}
}

@article{dpo,
  title={Direct preference optimization: Your language model is secretly a reward model},
  author={Rafailov, Rafael and Sharma, Archit and Mitchell, Eric and Manning, Christopher D and Ermon, Stefano and Finn, Chelsea},
  journal={Advances in neural information processing systems},
  volume={36},
  pages={53728--53741},
  year={2023}
}

@article{cheng2026dspark,
  title={DSpark: Confidence-Scheduled Speculative Decoding with Semi-Autoregressive Generation},
  author={Cheng, Xin and Yu, Xingkai and Shao, Chenze and Li, Jiashi and Xiong, Yunfan and Qian, Yi and Zhu, Jiaqi and Ma, Shirong and Zhang, Xiaokang and Ye, Jiasheng and others},
  journal={arXiv preprint arXiv:2607.05147},
  year={2026}
}

@misc{glm5team2026glm5vibecodingagentic,
      title={GLM-5: from Vibe Coding to Agentic Engineering},
      author={GLM-5-Team and : and Aohan Zeng and Xin Lv and Zhenyu Hou and Zhengxiao Du and Qinkai Zheng and Bin Chen and Da Yin and Chendi Ge and Chenghua Huang and Chengxing Xie and Chenzheng Zhu and Congfeng Yin and Cunxiang Wang and Gengzheng Pan and Hao Zeng and Haoke Zhang and Haoran Wang and Huilong Chen and Jiajie Zhang and Jian Jiao and Jiaqi Guo and Jingsen Wang and Jingzhao Du and Jinzhu Wu and Kedong Wang and Lei Li and Lin Fan and Lucen Zhong and Mingdao Liu and Mingming Zhao and Pengfan Du and Qian Dong and Rui Lu and Shuang-Li and Shulin Cao and Song Liu and Ting Jiang and Xiaodong Chen and Xiaohan Zhang and Xuancheng Huang and Xuezhen Dong and Yabo Xu and Yao Wei and Yifan An and Yilin Niu and Yitong Zhu and Yuanhao Wen and Yukuo Cen and Yushi Bai and Zhongpei Qiao and Zihan Wang and Zikang Wang and Zilin Zhu and Ziqiang Liu and Zixuan Li and Bojie Wang and Bosi Wen and Can Huang and Changpeng Cai and Chao Yu and Chen Li and Chengwei Hu and Chenhui Zhang and Dan Zhang and Daoyan Lin and Dayong Yang and Di Wang and Ding Ai and Erle Zhu and Fangzhou Yi and Feiyu Chen and Guohong Wen and Hailong Sun and Haisha Zhao and Haiyi Hu and Hanchen Zhang and Hanrui Liu and Hanyu Zhang and Hao Peng and Hao Tai and Haobo Zhang and He Liu and Hongwei Wang and Hongxi Yan and Hongyu Ge and Huan Liu and Huanpeng Chu and Jia'ni Zhao and Jiachen Wang and Jiajing Zhao and Jiamin Ren and Jiapeng Wang and Jiaxin Zhang and Jiayi Gui and Jiayue Zhao and Jijie Li and Jing An and Jing Li and Jingwei Yuan and Jinhua Du and Jinxin Liu and Junkai Zhi and Junwen Duan and Kaiyue Zhou and Kangjian Wei and Ke Wang and Keyun Luo and Laiqiang Zhang and Leigang Sha and Liang Xu and Lindong Wu and Lintao Ding and Lu Chen and Minghao Li and Nianyi Lin and Pan Ta and Qiang Zou and Rongjun Song and Ruiqi Yang and Shangqing Tu and Shangtong Yang and Shaoxiang Wu and Shengyan Zhang and Shijie Li and Shuang Li and Shuyi Fan and Wei Qin and Wei Tian and Weining Zhang and Wenbo Yu and Wenjie Liang and Xiang Kuang and Xiangmeng Cheng and Xiangyang Li and Xiaoquan Yan and Xiaowei Hu and Xiaoying Ling and Xing Fan and Xingye Xia and Xinyuan Zhang and Xinze Zhang and Xirui Pan and Xu Zou and Xunkai Zhang and Yadi Liu and Yandong Wu and Yanfu Li and Yidong Wang and Yifan Zhu and Yijun Tan and Yilin Zhou and Yiming Pan and Ying Zhang and Yinpei Su and Yipeng Geng and Yong Yan and Yonglin Tan and Yuean Bi and Yuhan Shen and Yuhao Yang and Yujiang Li and Yunan Liu and Yunqing Wang and Yuntao Li and Yurong Wu and Yutao Zhang and Yuxi Duan and Yuxuan Zhang and Zezhen Liu and Zhengtao Jiang and Zhenhe Yan and Zheyu Zhang and Zhixiang Wei and Zhuo Chen and Zhuoer Feng and Zijun Yao and Ziwei Chai and Ziyuan Wang and Zuzhou Zhang and Bin Xu and Minlie Huang and Hongning Wang and Juanzi Li and Yuxiao Dong and Jie Tang},
      year={2026},
      eprint={2602.15763},
      archivePrefix={arXiv},
      primaryClass={cs.LG},
      url={https://arxiv.org/abs/2602.15763},
}

@article{xu2026deepseek,
  title={Deepseek-v4. 1-flash: Pushing the limits of kv cache compression},
  author={Xu, Anyi and Li, B and Lin, Bangcai and Xue, Bing and Xian, BingCheng and Xu, Bingzheng and Wu, Bochao and Zhang, Bowei and Deng, Boyi and Yu, CC and others},
  journal={arXiv preprint arXiv:2609.19969},
  year={2026}
}

@inproceedings{medusa,
author = {Cai, Tianle and Li, Yuhong and Geng, Zhengyang and Peng, Hongwu and Lee, Jason D. and Chen, Deming and Dao, Tri},
title = {MEDUSA: Simple LLM inference acceleration framework with multiple decoding heads},
year = {2024},
publisher = {JMLR.org},
booktitle = {Proceedings of the 41st International Conference on Machine Learning},
articleno = {203},
numpages = {27},
location = {Vienna, Austria},
series = {ICML'24}
}

@inproceedings{eagle3,
 author = {Li, Yuhui and Wei, Fangyun and Zhang, Chao and Zhang, Hongyang},
 booktitle = {Advances in Neural Information Processing Systems},
 doi = {10.52202/085713-4562},
 editor = {D. Belgrave and C. Zhang and H. Lin and R. Pascanu and P. Koniusz and M. Ghassemi and N. Chen},
 pages = {136737--136756},
 publisher = {Curran Associates, Inc.},
 title = {EAGLE-3: Scaling up Inference Acceleration of Large Language Models via Training-Time Test},
 volume = {38, Main Conference},
 year = {2025}
}

@inproceedings{distillspec,
 author = {Zhou, Yongchao and Lyu, Kaifeng and Rawat, Ankit Singh and Menon, Aditya Krishna and Rostamizadeh, Afshin and Kumar, Sanjiv and Kagy, Jean-Fran\c{c}ois and Agarwal, Rishabh},
 booktitle = {International Conference on Learning Representations},
 editor = {B. Kim and Y. Yue and S. Chaudhuri and K. Fragkiadaki and M. Khan and Y. Sun},
 pages = {32011--32050},
 title = {DistillSpec: Improving Speculative Decoding via Knowledge Distillation},
 volume = {2024},
 year = {2024}
}

@inproceedings{judgedecoding,
 author = {Bachmann, Gregor and Anagnostidis, Sotiris and Pumarola, Albert and Georgopoulos, Markos and Sanakoyeu, Artsiom and Du, Yuming and Schoenfeld, Edgar and Thabet, Ali and Kohler, Jonas},
 booktitle = {International Conference on Learning Representations},
 editor = {Y. Yue and A. Garg and N. Peng and F. Sha and R. Yu},
 pages = {40894--40914},
 title = {Judge Decoding: Faster Speculative Sampling Requires Going Beyond Model Alignment},

 volume = {2025},
 year = {2025}
}

@article{yuan2025nondeterminism,
  title={Understanding and mitigating numerical sources of nondeterminism in llm inference},
  author={Yuan, Jiayi and Li, Hao and Ding, Xinheng and Xie, Wenya and Li, Yu-Jhe and Zhao, Wentian and Wan, Kun and Shi, Jing and Hu, Xia and Liu, Zirui},
  journal={Advances in Neural Information Processing Systems},
  volume={38},
  pages={169819--169851},
  year={2025}
}

@article{he2025nondeterminism,
  author = {Horace He and Thinking Machines Lab},
  title = {Defeating Nondeterminism in LLM Inference},
  journal = {Thinking Machines Lab: Connectionism},
  year = {2025},
  note = {https://thinkingmachines.ai/blog/defeating-nondeterminism-in-llm-inference/},
  doi = {10.64434/tml.20250910}
}

@InProceedings{kivi,
  title = 	 {{KIVI}: A Tuning-Free Asymmetric 2bit Quantization for {KV} Cache},
  author =       {Liu, Zirui and Yuan, Jiayi and Jin, Hongye and Zhong, Shaochen and Xu, Zhaozhuo and Braverman, Vladimir and Chen, Beidi and Hu, Xia},
  booktitle = 	 {Proceedings of the 41st International Conference on Machine Learning},
  pages = 	 {32332--32344},
  year = 	 {2024},
  editor = 	 {Salakhutdinov, Ruslan and Kolter, Zico and Heller, Katherine and Weller, Adrian and Oliver, Nuria and Scarlett, Jonathan and Berkenkamp, Felix},
  volume = 	 {235},
  series = 	 {Proceedings of Machine Learning Research},
  month = 	 {21--27 Jul},
  publisher =    {PMLR},
  url = 	 {https://proceedings.mlr.press/v235/liu24bz.html}
}

@article{deepseekv32,
  title={Deepseek-v3. 2: Pushing the frontier of open large language models},
  author={Liu, Aixin and Mei, Aoxue and Lin, Bangcai and Xue, Bing and Wang, Bingxuan and Xu, Bingzheng and Wu, Bochao and Zhang, Bowei and Lin, Chaofan and Dong, Chen and others},
  journal={arXiv preprint arXiv:2512.02556},
  year={2025}
}

@inproceedings{
swebench,
title={{SWE}-bench: Can Language Models Resolve Real-world Github Issues?},
author={Carlos E Jimenez and John Yang and Alexander Wettig and Shunyu Yao and Kexin Pei and Ofir Press and Karthik R Narasimhan},
booktitle={The Twelfth International Conference on Learning Representations},
year={2024},
url={https://openreview.net/forum?id=VTF8yNQM66}
}

@inproceedings{sinha2025illusion,
  title={The illusion of diminishing returns: Measuring long horizon execution in llms},
  author={Sinha, Akshit and Arun, Arvindh and Goel, Shashwat and Staab, Steffen and Geiping, Jonas},
  booktitle={International Conference on Learning Representations},
  volume={2026},
  pages={35184--35211},
  year={2026}
}

@InProceedings{lin2025critical,
  title = 	 {Critical Tokens Matter: Token-Level Contrastive Estimation Enhances {LLM}’s Reasoning Capability},
  author =       {Lin, Zicheng and Liang, Tian and Xu, Jiahao and Liu, Qiuzhi and Wang, Xing and Luo, Ruilin and Shi, Chufan and Li, Siheng and Yang, Yujiu and Tu, Zhaopeng},
  booktitle = 	 {Proceedings of the 42nd International Conference on Machine Learning},
  pages = 	 {37906--37918},
  year = 	 {2025},
  editor = 	 {Singh, Aarti and Fazel, Maryam and Hsu, Daniel and Lacoste-Julien, Simon and Berkenkamp, Felix and Maharaj, Tegan and Wagstaff, Kiri and Zhu, Jerry},
  volume = 	 {267},
  series = 	 {Proceedings of Machine Learning Research},
  month = 	 {13--19 Jul},
  publisher =    {PMLR},
  url = 	 {https://proceedings.mlr.press/v267/lin25j.html}
}

@article{wang2025beyond,
  title={Beyond the 80/20 rule: High-entropy minority tokens drive effective reinforcement learning for llm reasoning},
  author={Wang, Shenzhi and Yu, Le and Gao, Chang and Zheng, Chujie and Liu, Shixuan and Lu, Rui and Dang, Kai and Chen, Xiong-Hui and Yang, Jianxin and Zhang, Zhenru and others},
  journal={Advances in Neural Information Processing Systems},
  volume={38},
  pages={115452--115486},
  year={2025}
}

@misc{datacurve2026deepswev11,
  title  = {DeepSWE v1.1: a cleaner, more reproducible benchmark for frontier coding agents},
  author = {Wenqi Huang and Peter Jiang},
  year   = {2026},
  url    = {https://github.com/datacurve-ai/deep-swe},
}

@article{zheng2024sglang,
  title={Sglang: Efficient execution of structured language model programs},
  author={Zheng, Lianmin and Yin, Liangsheng and Xie, Zhiqiang and Sun, Chuyue and Huang, Jeff and Yu, Cody H and Cao, Shiyi and Kozyrakis, Christos and Stoica, Ion and Gonzalez, Joseph E and others},
  journal={Advances in neural information processing systems},
  volume={37},
  pages={62557--62583},
  year={2024}
}

@inproceedings{sfad,
  title={SFAD: Speculative Factuality-Aware Decoding},
  author={Chen, Guanqiao and Wang, Di and Hu, Lijie},
  booktitle={Findings of the Association for Computational Linguistics: EMNLP 2026},
  pages={24221--24239},
  year={2026}
}

@article{DBLP:journals/corr/abs-2406-18629,
  publtype={informal},
  author={Xin Lai and Zhuotao Tian and Yukang Chen and Senqiao Yang and Xiangru Peng and Jiaya Jia},
  title={Step-DPO: Step-wise Preference Optimization for Long-chain Reasoning of LLMs},
  year={2024},
  cdate={1704067200000},
  journal={CoRR},
  volume={abs/2406.18629},
  url={https://doi.org/10.48550/arXiv.2406.18629}
}

@article{nl2repo,
  publtype={informal},
  author={Jingzhe Ding and Shengda Long and Changxin Pu and Huan Zhou and Hongwan Gao and Xiang Gao and Chao He and Yue Hou and Fei Hu and Zhaojian Li and Weiran Shi and Zaiyuan Wang and Daoguang Zan and Chenchen Zhang and Xiaoxu Zhang and Qizhi Chen and Xianfu Cheng and Bo Deng and Qingshui Gu and Kai Hua and Juntao Lin and Pai Liu and Mingchen Li and Xuanguang Pan and Zifan Peng and Yujia Qin and Yong Shan and Zhewen Tan and Weihao Xie and Zihan Wang and Yishuo Yuan and Jiayu Zhang and Enduo Zhao and Yunfei Zhao and He Zhu and Chenyang Zou and Ming Ding and Jianpeng Jiao and Jiaheng Liu and Minghao Liu and Qian Liu and Chongyao Tao and Jian Yang and Tong Yang and Zhaoxiang Zhang and Xinjie Chen and Wenhao Huang and Ge Zhang},
  title={NL2Repo-Bench: Towards Long-Horizon Repository Generation Evaluation of Coding Agents},
  year={2025},
  month={December},
  cdate={1764547200000},
  journal={CoRR},
  volume={abs/2512.12730},
  url={https://doi.org/10.48550/arXiv.2512.12730}
}

@misc{li2026agentbatonefficientmodelcollaboration,
      title={AgentBaton: Efficient Model Collaboration in Agentic Tasks}, 
      author={Junlong Li and Jindong Jiang and Yonggan Fu and Jiarui Yao and Hanrong Ye and Wonmin Byeon and Peter Belcak and Hao Zhang and Pavlo Molchanov},
      year={2026},
      eprint={2610.35959},
      archivePrefix={arXiv},
      primaryClass={cs.MA},
      url={https://arxiv.org/abs/2610.35959}, 
}
\bibliographystyle{mlsys2025}

\end{document}